\documentclass[letterpaper, 10 pt, conference]{ieeeconf}  % Comment this line out if you need a4paper

\IEEEoverridecommandlockouts                              % This command is only needed if 
\usepackage{graphics} % for pdf, bitmapped graphics files
\usepackage{graphicx}
\usepackage{epsfig} % for postscript graphics files
\usepackage{mathptmx} % assumes new font selection scheme installed
\usepackage{times} % assumes new font selection scheme installed
\usepackage{amsmath} % assumes amsmath package installed
\usepackage{xcolor}
\usepackage{bm}
\usepackage{amssymb}  % assumes amsmath package installed
\usepackage{fmtcount}
\usepackage{mathrsfs}
\usepackage{booktabs}
\usepackage{diagbox}  % 引入 diagbox 宏包
\usepackage{siunitx}
\usepackage{subfigure}
\usepackage{pgfplots}
\pgfplotsset{compat=1.18}

\definecolor{niceblue}{RGB}{30,144,255}
\newcommand{\figref}[1]{\mbox{Fig.~\ref{#1}}}
\newcommand\norm[1]{\left\lVert#1\right\rVert}
\DeclareMathAlphabet{\mathcal}{OMS}{cmsy}{m}{n}

\title{\LARGE \bf
Bipedal Robust Walking on Uneven Footholds: Piecewise Slope LIPM with Discrete Model Predictive Control
}

\author{Yapeng Shi$^{1*}$, Sishu Li$^{1, 2*}$, Yongqiang Wu$^{1}$, Junjie Liu$^{1}$, Xiaokun Leng$^{1}$, Xizhe Zang$^{2}$ and Songhao Piao$^{1}$% <-this % stops a space
\thanks{*These authors contributed equally to this work. This work was supported by the National Natural Science Foundation of China under Grand 52305072, the China Postdoctoral Science Foundation under Grand 2024M754189, Heilongjiang Provincial Postdoctoral Research Project under Grand LBH-Z24016, Joint-Guided Project of Heilongjiang Natural Science Foundation under Grand LH2024F028, and Shenzhen Special Fund for Future Industrial Development under Grand KJZD20230923114222045. (\textit{Corresponding author: Xiaokun Leng}.) }% <-this % stops a space
\thanks{$^{1}$Yapeng Shi, Sishu Li, Yongqiang Wu, Junjie Liu, Xiaokun Leng, and Songhao Piao are with the Faculty of Computing, Harbin Institute of Technology, Harbin, 150001, China {\tt\small shi.yapeng@hit.edu.cn, \{23s108219, 24s103252, 2021112637\}@stu.hit.edu.cn, \{lengxiaokun, piaosh\}@hit.edu.cn}}%
\thanks{$^{2}$Sishu Li and Xizhe Zang are with the State Key Laboratory of Robotics and System, Harbin Institute of Technology, Harbin, 150001, China 
{\tt\small zangxizhe@hit.edu.cn}}%
}

\begin{document}

\maketitle
\thispagestyle{empty}
\pagestyle{empty}

%%%%%%%%%%%%%%%%%%%%%%%%%%%%%%%%%%%%%%%%%%%%%%%%%%%%%%%%%%%%%%%%%%%%%%%%%%%%%%%%
\begin{abstract}

This study presents an enhanced theoretical formulation for bipedal hierarchical control frameworks under uneven terrain conditions. Specifically, owing to the inherent limitations of the Linear Inverted Pendulum Model (LIPM) in handling terrain elevation variations, we develop a Piecewise Slope LIPM (PS-LIPM). This innovative model enables dynamic adjustment of the Center of Mass (CoM) height to align with topographical undulations during single-step cycles. Another contribution is proposed a generalized Angular Momentum-based LIPM (G-ALIP) for CoM velocity compensation using Centroidal Angular Momentum (CAM) regulation. Building upon these advancements, we derive the DCM step-to-step dynamics for Model Predictive Control MPC formulation, enabling simultaneous optimization of step position and step duration. A hierarchical control framework integrating MPC with a Whole-Body Controller (WBC) is implemented for bipedal locomotion across uneven stepping stones. The results validate the efficacy of the proposed hierarchical control framework and the theoretical formulation.

\end{abstract}

\section{Introduction}
\begin{figure}[t] 
    \centering    
    \includegraphics[width=1\linewidth]{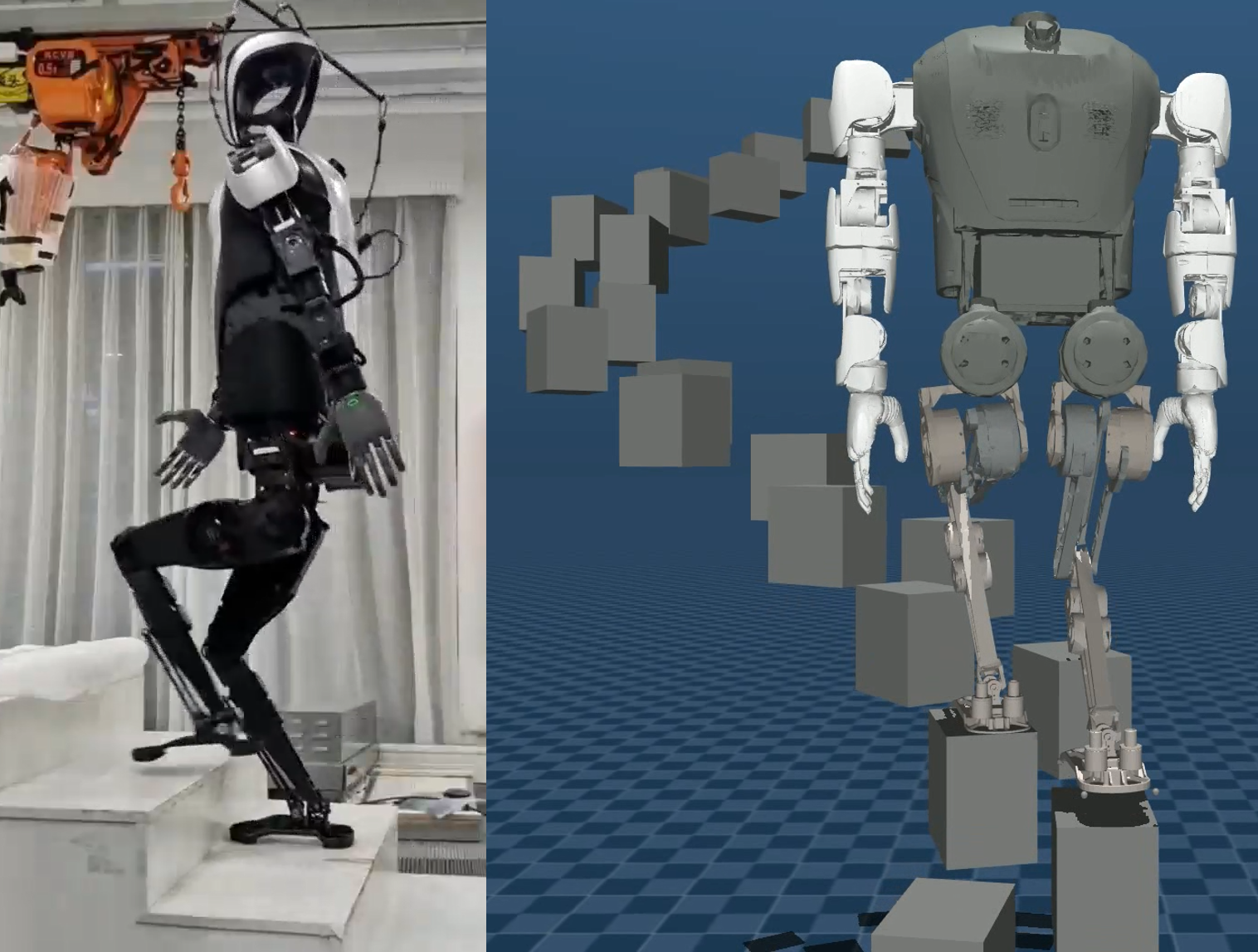}
    \caption{Humanoid robot KUAVO with its MuJoCo simulation walking on uneven stepping stones.}
    \label{steppingstones}
\end{figure}

Achieving adaptive bipedal locomotion across uneven terrains, particularly those with restricted footholds such as uneven stepping stones (Fig.~\ref{steppingstones}), remains a significant challenge in robotics. In such scenarios, the generation of walking gait patterns necessitates the synthesis among optimal step position, step duration, and Ground Reaction Forces (GRFs) to ensure stability \cite{khadiv2020walking}.

The Linear Inverted Pendulum Model (LIPM) has been extensively adopted as a computationally efficient paradigm for bipedal locomotion due to its simplified dynamics. By assuming a constant height for the Center of Mass (CoM), LIPM enables real-time motion control through linearized equations of motion \cite{kajita20013d}. This formulation has been further extended to accommodate inclined surfaces by aligning the CoM height relative to terrain slopes \cite{massah2012open}. However, a significant limitation persists: the model inherently excludes consideration of variable CoM height. With predefined step position and duration, the Center of Pressure (CoP) can be modified by carefully adjusting the joint torques, thereby modulating GRFs at the contact feet. However, exclusive reliance on CoP modification imposes critical limitations. This strategy not only requires specific physical leg designs but also limits flexibility in maintaining stability under unexpected terrain changes or external disturbances \cite{khadiv2020walking}.

Recent advancements highlight the superiority of adaptive step parameter modulation (step position and step duration) for stabilizing bipedal locomotion \cite{shi2019model, romualdi2022online}. The Divergent Component of Motion (DCM), which characterizes the unstable dynamics inherent in LIPM, provides a theoretical foundation for capturing and controlling the robot's dynamics stability during locomotion \cite{englsberger2013three}. DCM-based methodologies enable robots to dynamically adjust step positions, driving CoM trajectories toward stable manifolds, even under external disturbances or foothold constraints \cite{khadiv2016stepping, kamioka2018simultaneous}.

In recent years, the simultaneous adjustment of step position and step duration has been proposed using analytical \cite{castano2016dynamic} or optimization-based approaches \cite{kryczka2015online}. Model Predictive Control (MPC) frameworks have demonstrated particular promise by optimizing step parameters within a prediction horizon, accounting for system dynamics and foothold feasibility constraints \cite{xiang2024adaptive, bohorquez2017adaptive}. This enables the robot to generate reactive walking gait patterns that ensure stability over multiple steps. Complementing MPC, the Whole-Body Controller (WBC), on the other hand, operates at a lower level, enforces stability, handling contact dynamics, and ensuring the robot remains balanced and responsive to real-world conditions \cite{shi2022multi, chai2022survey}.

Dividing vertical and horizontal dynamics, LIPM has been used for horizontal motion planning in uneven terrain locomotion \cite{Dai2022bipeal}. Gibson \emph{et al.} proposes an Angular Momentum Linear Inverted Pendulum (ALIP) based controller that can walk on surfaces with varied inclinations \cite{Gibson2022terrain}. Oluwami \emph{et al.} introduces a variation of the ALIP model for walking stairs planning \cite{Oluwami2023stair}, demonstrates walking on periodic uneven discrete footholds. Walking on stochastic footholds in the sagittal plane has also been proposed \cite{Dai2022bipeal}. However, they neglect critical interactions between vertical centroid motion and horizontal dynamics when there are rapid changes in terrains with varying elevations. 
Building upon these foundations and limitations, this study proposes a novel extended LIPM considering coupled three-dimensional dynamics, aims to advance the state-of-the-art in bipedal locomotion on uneven terrains with discrete foothold constraints.

Our principal contribution comprises two novel dynamics models: 1) the Piecewise Slope LIPM (PS-LIPM), which enables variation in the CoM height during single-step cycles, and 2) the Generalized Angular-Momentum Linear Inverted Pendulum (G-ALIP) model, which facilitates CoM velocity regulation through adjustments of the Centroidal Angular Momentum (CAM).
The integration of these models establishes a DCM step-to-step dynamics, which explicitly characterizes hybrid dynamics during step transitions. As an additional contribution, we introduce an MPC-based optimization approach capable of simultaneously optimizing step position and step duration, allowing the bipedal to effectively traverse uneven terrains. To bridge the gap between reduced-order models and physical robots, we implement a hierarchical control framework that leverages a 1kHz WBC for real-time discrepancy compensation. This enhances robustness against terrain uncertainties and force disturbances. Experimental validation across a variety of challenging terrains and disturbance scenarios has confirmed the effectiveness of the framework. Comparative studies further demonstrate the superiority of PS-LIPM and G-ALIP formulations.

\section{Model Formulation}

\subsection{PS-LIPM Hybrid Dynamics}

\begin{figure}[tb]
    \centering
    \includegraphics[width=1\linewidth]{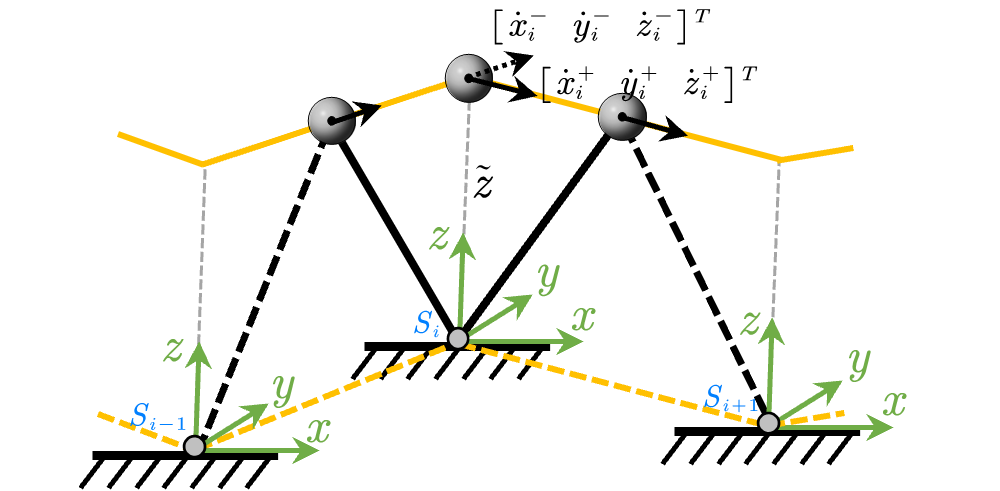}  
    \caption{As the contact point lands on the new virtual support plane (yellow dashed line), the CoM transitions from the old slope to the new slope (yellow solid line), with the velocity undergoing a sudden change at their intersection, noting that the two transitions do not occur simultaneously.} 
    \label{PS-LIPM}
\end{figure}

LIPM simplifies the complexity of bipedal dynamics into linear equations for real-time execution. This work extends the conventional LIPM to address height-varied terrains by piecewise-slope modeling of adjacent contact surfaces with different gradients (Fig.~\ref{PS-LIPM}). For the $x-z$ plane dynamics, the LIPM for slope surfaces is formulated as:
\begin{equation} \label{eq:lip_dynamics}
\begin{split}
\ddot{x}  &= \frac{g}{\tilde z } x  =: \omega^2 x  \\   
\tilde z  &= z  - k_x x   ,
\end{split}
\end{equation}
where $x$, $\dot{x}$, and $\ddot{x}$ are the position, velocity, and acceleration of the CoM, respectively, $g$ is the gravitational acceleration, $z$ indicates CoM height, $\tilde z $ specifies the vertical distance between CoM and slope surface, $k_x$ defines the slope gradient coefficient, subscript notation $(\cdot)_x$ for $x$ direction, $(\cdot)_y$ for $y$ direction, and $\omega = \sqrt{{g}/{\tilde z }}$ denotes the system's natural frequency. Without loss of generality, the current contact point $S=\left[S_x,S_y,S_z\right]^T$ is analytically set at $(0,0,0)$ to establish a local coordinate system.

The gradients of the piecewise slopes vary during the traversal of the stepping stones with random elevations. The robot must dynamically transition between piecewise slopes with distinct gradients. The following derivation is based on the assumption that velocity variations during slope transitions occur within an infinitesimal time interval, allowing them to be modeled as instantaneous state switching.

This study introduces the superscript notation $\left(\cdot\right)^-$ for pre-transition states and $\left(\cdot\right)^+$ for post-transition states to characterize dynamic switching between inclined surfaces. The CoM states defined correspondingly as: 1) Pre-transition: $ \left [x, y, z, \dot{x}^{-}, \dot{y}^{-}, \dot{z}^{-} \right ]^T $, 2) Post-transition: $ \left [ x, y, z, \dot{x} ^{+}, \dot{y} ^{+}, \dot{z}^{+} \right ]^T $. Slope characteristics are parameterized by gradient coefficients with $\left(k_{x}^{-}, k_{y}^{-} \right)$ and $\left(k_{x}^{+}, k_{y}^{+} \right)$, respectively. Thus, the states must satisfy:
\begin{equation} \label{eq:Vz_PSLIP}
\begin{split}
    \dot{z} ^{-} &=k_{x}^{-} \dot{x} ^{-}+k_{y}^{-} \dot{y} ^{-} \\
    \dot{z} ^{+} &=k_{x}^{+} \dot{x} ^{+}+k_{y}^{+} \dot{y} ^{+}   .
\end{split}
\end{equation}

Angular momentum conservation about the contact point yields:
\begin{equation} \label{eq:am_PSLIP}
\begin{split}
L_x: \ & m  \left(y  \dot{z}^{-} + z  \dot{y}^{-}  \right) = m  \left(y  \dot{z}^{+} + z  \dot{y}^{+}  \right) \\
L_y: \  & m  \left(z  \dot{x}^{-}  - x  \dot{z}^{-} \right) = m  \left(z  \dot{x}^{+}  - x  \dot{z}^{+} \right)   ,
\end{split}
\end{equation}
where $L_x, L_y$ denotes the components of angular momentum with respect to the contact point, $m$ denotes mass of the system.

By solving Eq.~\eqref{eq:am_PSLIP}, the following PS-LIPM dynamics can be derived:
\begin{equation} \label{eq:delta_xyz}
\begin{split}
\dot{x} ^{+} &= \frac{x }{z }  \Delta \dot{z} + \dot{x} ^{-} \\
\dot{y} ^{+} &= \frac{y }{z }  \Delta \dot{z} + \dot{y} ^{-} \\
\dot{z} ^{+} &= \dot{z} ^{-} + \Delta \dot{z}   \ .
\end{split} 
\end{equation}

According to Eq.\eqref{eq:Vz_PSLIP}, $\Delta \dot{z}$ satisfies:
\begin{equation}
\Delta \dot{z} =\frac{\left(k_{x}^{+}-k_{x}^{-}\right) \dot{x} ^{-}+\left(k_{y}^{+}-k_{y}^{-}\right) \dot{y} ^{-}}{1-k_{x}^{+} \frac{x }{z }-k_{y}^{+} \frac{y }{z }} .
\end{equation}

Let $\bm{x}$ denote the robot's state vector, formally defined as:
\begin{equation}
\bm{x}=
\left[x,y,\dot{x},\dot{y} \right]^T  \in \mathbb{R}^4  ,
\end{equation}
and $\Delta _{\mathrm{com}}$ is the corresponding reset map. Eq.~\eqref{eq:delta_xyz} can be rewritten to:
\begin{equation} \label{eq:delta_Xz}
\begin{split}
\bm{x}^+ &=\Delta _{\mathrm{com}}(\bm{x}^-) \\
\dot{z} ^{+} &= \dot{z} ^{-} + \Delta \dot{z}   \ .
\end{split} 
\end{equation}

To formally distinguish between boundary states at slope transitions and general system states, we define the following hybrid system domain:
\begin{equation}
\begin{array}{cc}
     \mathcal{D} = \left\{\bm{x} \in \mathbb{R}^4: z \neq k_x^+x+k_y^+y + \tilde{z} \right\}    
    \\
     \mathcal{S} = \left\{\bm{x} \in \mathbb{R}^4: z = k_x^+x+k_y^+y + \tilde{z} \right\}    .
\end{array}  
\end{equation}

The complete hybrid dynamics of the PS-LIPM are then derived by combining Eq.~\eqref{eq:lip_dynamics} and Eq.~\eqref{eq:delta_Xz}:
\begin{equation} 
\left\{
\begin{array}{l}
\left\{
\begin{array}{l}
\ddot{x} = \omega^2  x \\
\ddot{y} = \omega^2  y
\end{array}
\right.
\quad \bm{x} \in \mathcal{D/S}
\\
\left\{
\begin{array}{l}
\bm{x}^+ =\Delta _{\mathrm{com}}(\bm{x}^-)
\end{array}
\right.
\quad
\bm{x}^- \in \mathcal{S}  \ .
\end{array}
\right.
\end{equation}
where $\Delta _{\mathrm{com}}$ denotes the reset map introduced in Eq.~\eqref{eq:delta_xyz}.

\subsection{G-ALIP Model} 

The Angular-Momentum Linear Inverted Pendulum (ALIP) model extends the classic LIPM. ALIP model explicitly includes angular momentum about the contact point as a state variable. This critical enhancement enables more physically accurate representations of bipedal dynamics:
\begin{equation} \label{eq:ALIP}
\begin{bmatrix}
\dot{x} \\
\dot{L}_y
\end{bmatrix}=
\begin{bmatrix}
0 & 1/(mz) \\
mg & 0
\end{bmatrix}
\begin{bmatrix}
x \\
L_y
\end{bmatrix}\,.
\end{equation}

Given the definition $\tilde{\dot{x}}^{\scriptstyle\mathrm{ALIP}} =  L_y/ \left( mz\right) $, the Eq.~\ref{eq:ALIP} can be reformulated as: 
\begin{equation}
\begin{bmatrix}
\dot{x} \\
\tilde{\ddot{x}}^{\scriptstyle\mathrm{ALIP}}
\end{bmatrix}=
\begin{bmatrix}
0 & 1 \\
\omega^2 & 0
\end{bmatrix}
\begin{bmatrix}
x \\
\tilde{\dot{x}}^{\scriptstyle\mathrm{ALIP}}
\end{bmatrix}  \,.
\end{equation}

$L_y$ comprises both the angular momentum of the CoM about the contact point and the CAM $L_{y,\mathrm{com}}$:
\begin{equation}
    L_y = mz\dot{x}-mx\dot{z}+L_{y,\mathrm{com}}\,.
\end{equation}

For a robot walking on a planar surface, $\dot{z}$ can be assumed to be zero. Thus, the velocity term $\tilde{\dot{x}}^{\scriptstyle\mathrm{ALIP}}$ satisfies:
\begin{equation}
\tilde{\dot{x}}^{\scriptstyle\mathrm{ALIP}} = \dot{x} + L_{y,\mathrm{com}}/ \left( mz\right)  \,.
\end{equation}

This formular introduces a compensatory mechanism that converts the angular momentum to an equivalent CoM velocity when abstracting the robotic system to an inverted pendulum model. The equivalence implies instantaneous transformation of centroidal momentum from angular to linear form. In contrast, the conventional LIPM entirely neglects changes in CAM. However, the behavior of a physically realizable robot under appropriate postural control is expected to lie between these two models, with its posture gradually stabilizing over time and its CAM asymptotically approaching zero.

Therefore, we propose a G-ALIP model:
\begin{equation}
\tilde{\dot{x}} = \dot{x} + 
\alpha L_{y,\mathrm{com}}/ \left( mz\right)  ,\quad
\alpha  \in [0, 1]   \,.
\end{equation}

This model introduces a G-ALIP coefficient $\alpha$ to represent the conversion ratio between the CAM and linear velocity. $\alpha=0$ corresponds to the conventional LIPM, while $\alpha=1$ corresponds to that of ALIP model. In our work, by selecting an appropriate value for this coefficient between zero and one, the accuracy of the prediction can be improved.

\section{DCM Step-to-Step Dynamics}
This section derive the step-to-step dynamics of initial DCM under extended LIPM. We first derive the hybrid dynamics of the piecewise-slope DCM under the G-ALIP framework. Then we present the preliminaries of the LIPM nominal trajectory. The asymptotical convergence of the LIP trajectory under nominal DCM is proved. Finally, the step-to-step dynamics of the DCM based on the LIP nominal trajectory approximation is ultimately presented.

Regulating the DCM ensures dynamic balance maintenance throughout the robotic walking cycle. The DCM is defined as follows:
\begin{equation} \label{eq:DCM_defination}
\left\{\begin{array}{l}
\xi_{x}=x+\frac { \tilde{\dot{x}}}{\omega} \\
\xi_{y}=y+\frac { \tilde{\dot{y}}}{\omega}
\end{array}
\right.\,.
\end{equation}

By combining Eq.~\eqref{eq:lip_dynamics} and Eq.~\eqref{eq:DCM_defination}, we can obtain:
\begin{equation}
\left\{\begin{array}{l}\dot{\xi_x}=\omega\left(\xi_x-S_{x}\right) \\ \tilde{\dot{x}}=\omega(\xi_x-x)\end{array}\right.\,.
\end{equation}

The time evolution of the DCM is given by:
\begin{equation}
\xi_x(t)=\left(\xi_{x,0}-S_{x}\right) e^{\omega t}+S_{x}\,,
\end{equation}
where $\xi_{x,0}$ is the initial DCM at $t = 0$. From the above formula, it can be observed that $x$ asymptotically converges toward $\xi_x$, while $\xi_x$ itself demonstrates inherent exponential divergence characteristics. This divergence property necessitates careful stability management in bipedal locomotion. As $ t \rightarrow \infty $, the DCM exponentially diverges ($e^{\omega t}$ term). 

The hybrid dynamics formulation of the PS-LIPM reveals critical discontinuities in the CoM velocity during slope transitions. The intrinsic coupling between $\xi_x$ and the CoM velocity leads to discontinuous state transitions for the DCM. The DCM hybrid dynamics can be expressed as follows:
\begin{equation}
\left\{
\begin{array}{l}
\dot{\xi}_{x} =\omega\left(\xi_x-S_{x}\right) 
\quad\quad
\bm{x} \in \mathcal{D/S}
\\
\xi_x^{+}= \xi_x^{-}+ \frac{1}{\omega}\left(\tilde{\dot{x}}^{+}-\tilde{\dot{x}}^{-} \right) 
\quad
\bm{x}^- \in \mathcal{S}
\end{array}
\right.\,,
\end{equation}
such that $[x^-,y^-,\tilde{\dot{x}}^-,\tilde{\dot{y}}^-]^T = \Delta _{\mathrm{com}}\left([x^+,y^+,\tilde{\dot{x}}^+,\tilde{\dot{y}}^+]^T\right)$.

% \subsection{Nominal LIPM Trajectory of Periodic Gaits}

By solving Eq.~\eqref{eq:lip_dynamics}, the analytical solution of the CoM state can be derived as follows:
\begin{equation}
x{\left(t\right)}=x_{0} \cosh (\omega t)+\dot{x}_{0} \sinh (\omega t) / \omega \,.
\end{equation}
% where $t$ here is the elapsed time.

\begin{figure}[t]
    \centering
    \includegraphics[width=0.9\linewidth]
    {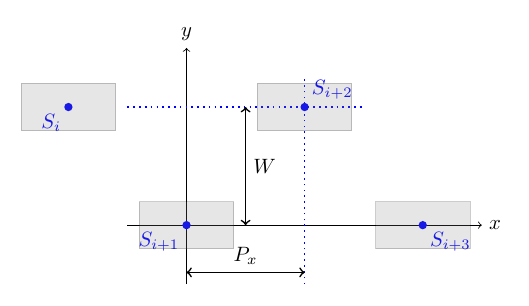}
    \caption{Step positions of the robot during periodic motion without lateral offset. The blue dots are the step positions and the gray rectangles represent the stepping stones.}
    \label{fig: placement_periodic}
\end{figure}

Consider a periodic motion trajectory, as depicted in \figref{fig: placement_periodic}. Let $P_x$ denote the step length in the forward direction, $W$ the lateral distance between the two feet, and $T$ the step duration. The subscript $(\cdot)_i$ indicates the step index, and the subscript $(\cdot)_m$ denotes the midpoint of each motion cycle. Defining the midpoint at $t_{m}=0$, the motion cycle spans the temporal domain $\left[-T/2,T/2 \right]$.

Due to symmetry, it is evident that $x_m=0$, $\dot{y}_m=0$ and the following relationship holds:
\begin{equation}
\begin{split}
    \dot{x}_m = \frac{\omega P_x}{2\text{sinh}(\omega T / 2)}   \\
    y_m = \frac{W}{2\text{cosh}(\omega T / 2)}
\end{split}\,.
\end{equation}

Therefore, the periodic trajectory can be mathematically formulated as:
\begin{equation}
\left\{
\begin{aligned}
    x(t) &= \dot{x}_m \sinh(\omega t)/\omega \\
    y(t) &=  y_m\cosh(\omega t) + \dot{y}_m\sinh(\omega t)/\omega
\end{aligned}
\right.\,.
\end{equation}

When introducing a lateral offset $P_y$ in the $y$-direction, $\dot{x}_m$, $x_m$ and $y_m$ remain invariant, whereas $\dot{y}_m$ is adjusted to $\omega P_y/\left(2\text{sinh}(\omega T/2)\right)$.

Under the assumption of periodic variations in both step duration and step position, the DCM exhibits periodic behavior, yielding the nominal DCM trajectory:
\begin{equation}
\boldsymbol{\xi}_{0, \mathrm{nom}}=\boldsymbol{\xi}_{0, \mathrm{nom}} - S_{\mathrm{init}}=\left[\begin{array}{c}\frac{P_x}{e^{\omega T}-1} \\ \frac{W_{l / r}}{e^{\omega T}+1}+\frac{Py}{e^{\omega T}-1}\end{array}\right]\,,
\end{equation}
where $S_{\mathrm{init}} = (0,0)^T$ is the initial contact point, $ W_{l / r} = -W$ when the supporting leg is the left leg, and $W_{l / r} = W$ otherwise, as shown in \cite{xiang2024adaptive}.

The asymptotic convergence of the DCM to its nominal trajectory guarantees the exponential stabilization of the centroid motion to periodic orbits. Let $x_{\mathrm{dev}}$ represent the initial CoM position deviation at $t = -T/2$. Through substitution of this perturbed initial condition into the governing dynamics, the relationship between the initial and final states of the cycle and the nominal DCM can be expressed as follows:
\begin{equation}
\left\{
\begin{aligned}
     &x_{-T/2} = -\frac{P_x}{2}+x_{\mathrm{dev}} \\
     &x_{-T/2}+ \dot{x}_{-T/2} = \frac{P_x}{e^{\omega T} -1} \\
     &x_{T/2} = x_{-T/2} \text{cosh}(\omega T) + \dot{x}_{-T/2}\text{sinh}(\omega T) /\omega
\end{aligned}
\right.\,.
\end{equation}

The CoM position at $x_{T/2}$ can be derived as follows:
\begin{equation}
    x_{T/2} = \frac{P_x}{2}+e^{-\omega T}x_{\mathrm{dev}}\,.
\end{equation}

As demonstrated in the derivation, the introduced initial CoM position deviation asymptotically diminishes to zero. This convergence behavior guarantees that the centroid motion ultimately stabilizes to the periodic trajectory.

The hybrid dynamics of the DCM govern the evolution of the robot's state during step durations through the coupled relationship between the DCM and centroid velocity. Given the convergence of centroid trajectories toward nominal periodic gait patterns, the actual centroid motion can be effectively approximated by its nominal trajectory. Furthermore, empirical observations reveal that slope transition timing typically coincides with the midpoint of the step cycle under normal operating conditions. These simplifications yield the following step-to-step DCM dynamics:
\begin{equation}\label{eq: S2S_DCM}
    \xi_{x,T/2} = \left(\xi_{x,-T/2} -S_x\right)e^{\omega T} + \frac{\dot{x}_m^+ - \dot{x}_m^- }{\omega}e^{\omega T /2} + S_x\,,
\end{equation}
such that $\bm{x}_{m}^+ = \Delta _{\mathrm{com}}\left( \bm{x}_{m}^-\right)$.

\section{Hierarchical Control Framework Based on DCM Step-to-Step Dynamics}

\begin{figure}[t]
    \centering
    \includegraphics[width=0.9\linewidth]
    {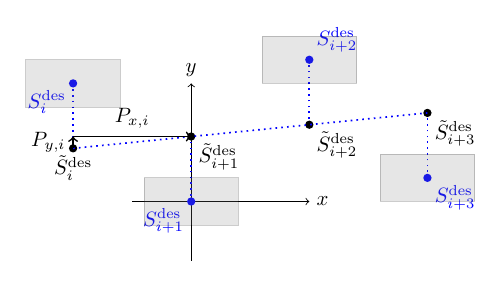}
    \caption{Segmented slope virtual constraint generation. The blue dots are the desired step positions and black dots are adjusted step positions used to generate virtual slope considering the fixed step width $W$.}
    \label{fig: virtual_slope}
    \vspace{-5mm}
\end{figure}

This study presents a hierarchical control framework, consisting of an MPC-based planner and a WBC, for bipedal locomotion on uneven terrains. The proposed framework abstracts discontinuous stepping-stone terrains as sequences of virtual slopes, with geometric parameters adaptively synthesized from spatial stone distributions. As shown in \figref{fig: virtual_slope}, considering the lateral foot distance $W$, we introduce an adjusted foothold
$\tilde{S}_i^{\mathrm{des}} = S_i^{\mathrm{des}} + \left[0, W_{\text{l / r},i}/2,0 \right]^T$. The adjustment enables proper virtual slope generation without direct connection of adjacent footholds. The virtual slope vector between consecutive adjusted footholds is as follows:
\begin{equation}
    \left[P_{x,i},P_{y,i},P_{z,i} \right]^T = \tilde{S}_{i+1}^{\mathrm{des}} -\tilde{S}_{i}^{\mathrm{des}} .
\end{equation}

The slope gradient is computed under the geometric constraint that the connection line between adjacent footholds remains perpendicular to the intersection line of the virtual slope plane with the $xy$-plane.

The slope gradient can be calculated as follows:
\begin{equation}
\begin{aligned}
    k_{x,i}^{+} &= k_{x,i+1}^{-} = \frac{P_{x,i}  P_{z,i}}{{P_{x,i}^2+P_{z,i}^2}} \\
    k_{y,i}^{+} &= k_{y,i+1}^{-} = \frac{P_{y,i}  P_{z,i}}{{P_{y,i}^2+P_{z,i}^2}}
\end{aligned}\,.
\end{equation}

The high-level gait planner incorporates the methodology mentioned in \cite{xiang2024adaptive} for trajectory optimization.
To formulate the trajectory optimization problem, we define the key variables: 
\begin{itemize}
    \item $\tau_{i}=e^{\omega T_{i}}$, represents the temporal variable.
    \item $\mathbf{b}_{i} = [\xi_{x,i} - S_{x,i},\xi_{y,i}-S_{y,i}]^T$, corresponds to the DCM offset.
    \item $\mathbf{u}_i = [S_{x,i} - S_{x,0},S_{y,i}-S_{y,0}]^T$, redefined as the displacement vector from the $i$-th step position to its current contact point $S_0$, rather than the previous step position. In particular, $\mathbf{u}_0 = [0,0]^T$. 
\end{itemize}

With the aforementioned definitions of symbols, by substituting $\xi_{x,i-1}$ and ${\xi_{x,i}}$ into  Eq.~\eqref{eq: S2S_DCM}, we can derive the solution as follows:
\begin{equation}
\tau_{i}\mathbf{b}_{i-1}-\mathbf{b}_{i}=\mathbf{u}_{i}-\mathbf{u}_{i-1}+
\left[\begin{array}{c}\frac{\dot{x}_{m,i}^+ - \dot{x}_{m,i}^-}{\omega} \\ \frac{\dot{y}_{m,i}^+ - \dot{y}_{m,i}^-}{\omega}\end{array}\right] e^{\omega T_i /2}.
\end{equation}

We aim to make the aforementioned parameters as close as possible to the desired values:
\begin{itemize}
    \item $\tau_{i,{\mathrm{nom}}}=e^{\omega T_{i,\mathrm{nom}}}$, $T_{i,\mathrm{nom}}=$ \SI{0.5}{s}.
    \item $\mathbf{b}_{i, \text{nom}} = \boldsymbol{\xi}_{0, \mathrm{nom}}$, this desired value is for maintaining balance.
    \item $\mathbf{u}_{i, \text{nom}} = \left[S_{x,i}^{\mathrm{des}} - S_{x,0},S_{y,i}^{\mathrm{des}} - S_{y,0}\right]^T$, the center of each stepping stone is designated as the desired step position.
\end{itemize}

The discrete MPC problem is formulated as follows:
\begin{equation}
\begin{aligned}
\underset{\substack{\tau_{i}, \mathbf{b}_{i},\mathbf{u}_{i}}}{\mathrm{arg\ min}} \quad \quad
& \sum_{i=1}^{N} w_{\tau, i} \left\| \tau_{i} - \tau_{i, \text{nom}} \right\|^2 \\
+ & \sum_{i=1}^{N} w_{b, i} \left\| \mathbf{b}_{i} - \mathbf{b}_{i, \text{nom}} \right\|^2 \\
+ & \sum_{i=1}^{N} w_{u, i} \left\| \mathbf{u}_{i} - \mathbf{u}_{i, \text{nom}} \right\|^2   \  ,
\\
      \text{s.t.} \quad \tau_{i}\mathbf{b}_{i-1}&-\mathbf{b}_{i}=\mathbf{u}_{i}-\mathbf{u}_{i-1} + \frac{\Delta{\dot{\bm{v}}_{m,i}}}{\omega}e^{\omega T_{i,\mathrm{nom}} /2}  \,,
\end{aligned}
\end{equation}
where $\Delta \dot{\bm{v}}_{m,i} = \left[\dot{x}_{m,i}^+ - \dot{x}_{m,i}^-,\dot{y}_{m,i}^+ - \dot{y}_{m,i}^-\right]^T$, $w_{\tau, i}$, $w_{b, i}$, $w_{u, i}$ are weight coefficients. In order to minimize the discrepancy between the actual step position and the desired step position, the weight $w_{u, i}$ is set to a relatively large value. Conversely, the step duration, being a more flexible parameter and primarily subject to adjustment, is associated with a smaller weight $w_{\tau, i}$. 
In addition, the subscript $(\cdot)_\mathrm{nom}$ denotes the desired nominal value. We made an approximation in the constraint, where $T_i$ in the last term on the right-hand side of the equation was replaced by $T_{i,\text{nom}}$.

To bridge the divergence between simplified model and the physical system, we implement a WBC strategy proposed in \cite{shen2022implementation} as a low-level controller. This controller simultaneously tracks the desired centroidal trajectory and the foot-end trajectories generated by the MPC planner. The optimization problem is formulated as a quadratic programming(QP) problem with prioritized task execution.
The WBC implemented here extends the baseline WBC framework \cite{shen2022implementation} through an arm regularization task. The task aims to drive the arm toward a desired joint angle by introducing a cost term:
\begin{equation}
w_{\mathrm{arm}} \norm{\bm{\ddot{q}}_{\mathrm{arm}}-\bm{\ddot{q}}_{\mathrm{arm}}^{\mathrm{des}}}^2
\end{equation}

The desired acceleration is computed via a PD control law:
\begin{equation}
\bm{\ddot{q}}_{\mathrm{arm}}^{\mathrm{des}} = \bm{K}_p(\bm{q}_{\mathrm{arm}}^{\mathrm{des}} -\bm{q}_{\mathrm{arm}})+\bm{K}_d(\bm{\dot{q}}_{\mathrm{arm}}^{\mathrm{des}} -\bm{\dot{q}}_{\mathrm{arm}})
\end{equation}
where $\bm{q}_{\mathrm{arm}}$, $\bm{\dot{q}}_{\mathrm{arm}}$, and $\bm{\ddot{q}}_{\mathrm{arm}}$ denote the joint angle, joint velocity, and joint acceleration vector, respectively, and $(\cdot)^{\mathrm{des}}$ describes the desired state.
The controller assumes sufficient actuator torque capacity to execute generated motions, explicitly formulated through the omission of torque constraints in the QP formulation. This simplifying assumption facilitates controller synthesis while acknowledging potential implementation limitations through torque saturation effects in practical applications. 

\section{Results}
This section demonstrates the capability of the proposed hierarchical control framework in generating adaptive bipedal locomotion. The motion generation is achieved through simultaneous optimization of step duration and step position across stepping stones with elevation variations. When integrated with the WBC, the framework exhibits robust performance against terrain irregularities and external disturbances. Our tests were conducted in the physics-based simulation in MuJoCo \cite{todorov2012mujoco} at 1KHz control frequency. Experimental validation was conducted on KUAVO (Version 4.0), a newly developed humanoid robot featuring upgraded actuation and sensing modules (\figref{steppingstones}).
The detailed specifications related to this work is illustrated in Table \ref{table:specifications}. Normal height here is the distance from the hip yaw joint to the ground when KUAVO is standing normally. The size of the stones in all tests is \SI{0.2}{m} by \SI{0.14}{m}.

\begin{table}[t]
	\caption{Kuavo Specifications}
	\label{table:specifications}
	\vspace{-4mm}
	\begin{center}
		\begin{tabular}{|c||c||c||c|}
			\hline
			\bfseries Mass & \bfseries Normal height & \bfseries $W$ & \bfseries Foot size \\
			\hline
				\SI{44.9}{kg} & \SI{0.78}{m} & \SI{0.2}{m} & \SI{0.2}{m} $\times$ \SI{0.1}{m} \\
			\hline
		\end{tabular}
	\end{center}
	\vspace{-8mm}
\end{table}

We first established the baseline relative positions $\bm{p}^{\mathrm{init}}$ between adjacent stepping stones. Subsequently, we introduced terrain disturbance $\bm{p}^{\mathrm{dist}}$ in all three spatial directions, along with angular perturbations $\theta$ in the $z$-direction, to generate the actual stepping stones. Considering the spacing between the robot's left and right feet, the additional  offset $W_{l / r}$ is added and $W_{l / r}=\pm W$. The detailed terrain information used in the experiments is summarized in Table \ref{table:settings}. Uniformly distributed random values are represented using $U[\cdot,\cdot]$.

\begin{table}[b]
	\caption{Experiment Scenarios}
	\label{table:settings}
	\vspace{-6mm}
	\begin{center}
		\begin{tabular}{|c||c||c||c|}
			\hline
			\bfseries  & \bfseries $\bm{p}^{\mathrm{init}}$ (cm) & \bfseries $\theta$ (rad) & \bfseries $\bm{p}^{\mathrm{dist}}$ (cm) \\
			\hline
			\bfseries a & [20, 0, 0] & 0.0 & periodic $z$=$\pm17$  \\
                \hline
                \bfseries b & [20, 0, 10] & 0.2/-0.2 &  $U$([-2.5,2.5]×[-2.5,2.5]×[-5,5]) \\
			\hline
                \bfseries c & [20, 0, 0] & 0 & [0, 0, 0]  \\
			\hline      
		\end{tabular}
	\end{center}
\end{table}

\begin{figure*}[t!]
    \subfigure[KUAVO navigates a series of periodically arranged  stepping stones, with neighboring stones spaced \SI{\pm17}{cm} apart in the $z$ direction.]{ 
    \begin{minipage}[t][6cm][c]{0.48\linewidth} 
    \centering
    \includegraphics[width=\linewidth, height=5.5cm, keepaspectratio]{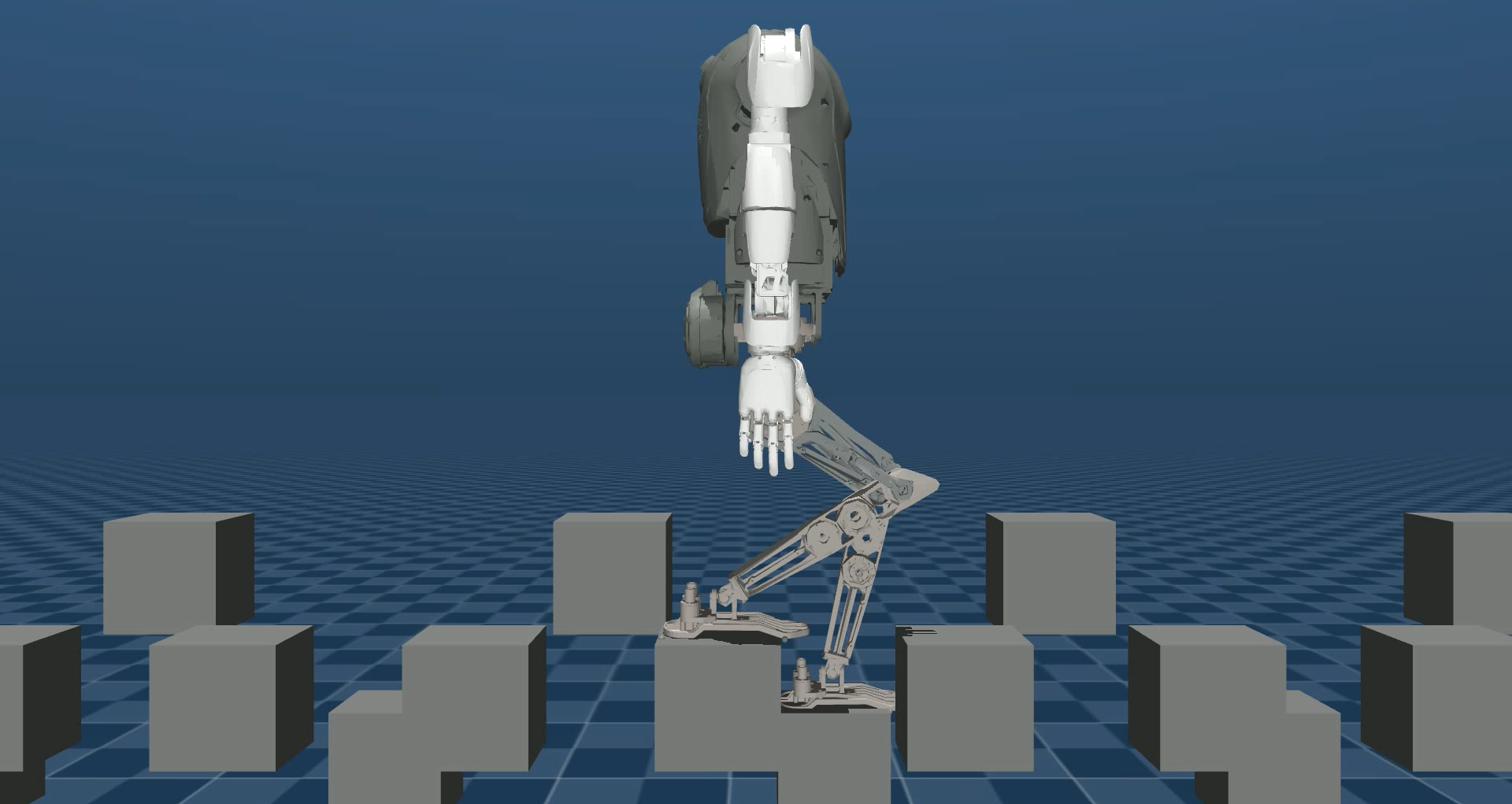}
    \label{fig:scenea_a}
    \end{minipage}
    }
    \hfill
    \subfigure[The blue dashed line and orange dashed line indicate the CoM trajectory and piecewise virtual slopes projected in the $xz$ plane.]{
    \begin{minipage}[t][6cm][c]{0.46\linewidth}
    \centering
    \includegraphics[width=\linewidth, height=6cm, keepaspectratio]{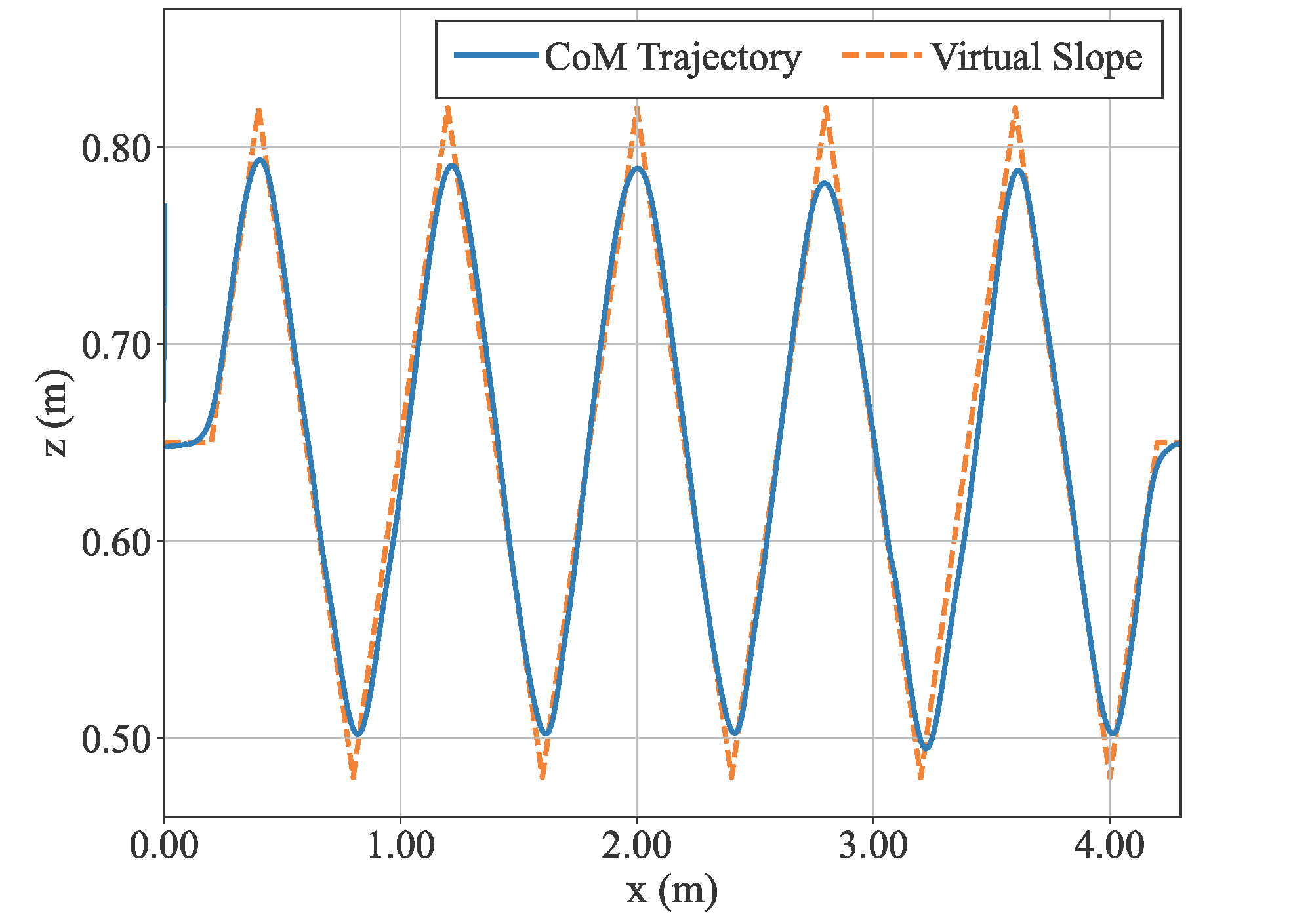}
    \label{fig:scenea_b}
    \end{minipage}
    }
\caption{KUAVO traverses over the stepping stones with periodic elevation.}
\label{fig:scenea}
\end{figure*}

We first validated the proposed framework across diverse terrains and operational conditions, followed by controlled comparative studies to evaluate the theoretical formulations of the proposed model in Section II and III. 
To balance the planner's performance and solution computational, the prediction horizon was set to a full stride duration, such that the MPC's prediction encompassed both left and right stepping phases within a complete gait cycle. The optimization problem was formulated using the Drake library \cite{drake}.
All tests are performed on a 12th Gen Intel i7 processor with 64GB of RAM. In this work, the average solver time is \SI{0.9784}{ms} for MPC and \SI{0.5881}{ms} for WBC, including the condition with terrain unevenness and force disturbances. 

In the first validation, the proposed approach was employed in two different scenarios and then evaluate its robustness under unexpected force disturbances. Detailed experimental settings are shown in Table \ref{table:settings}.

\textbf{Scenario a:} To explicitly demonstrate the CoM trajectory generated by the proposed MPC-based optimization architecture, the KUAVO humanoid robot was programmed to traverse periodic stepping stones with a periodic vertical offset of \SI{17}{cm} between adjacent stones (Fig.~\ref{fig:scenea}). 
As illustrated in Fig.~\ref{fig:scenea_b}, even when navigating uneven terrain with varying elevations, the CoM trajectory remains closely aligned with the direction of the constructed piecewise slopes. Notably, in the absence of PS-LIPM compensation or ALIP parameter adjustments, the robot failed to stabilize on this terrain and collapsed within $40^{th}$ step, highlighting the critical role of the proposed method.

\begin{figure}[h!]
    \centering
    \includegraphics[width=1\linewidth]{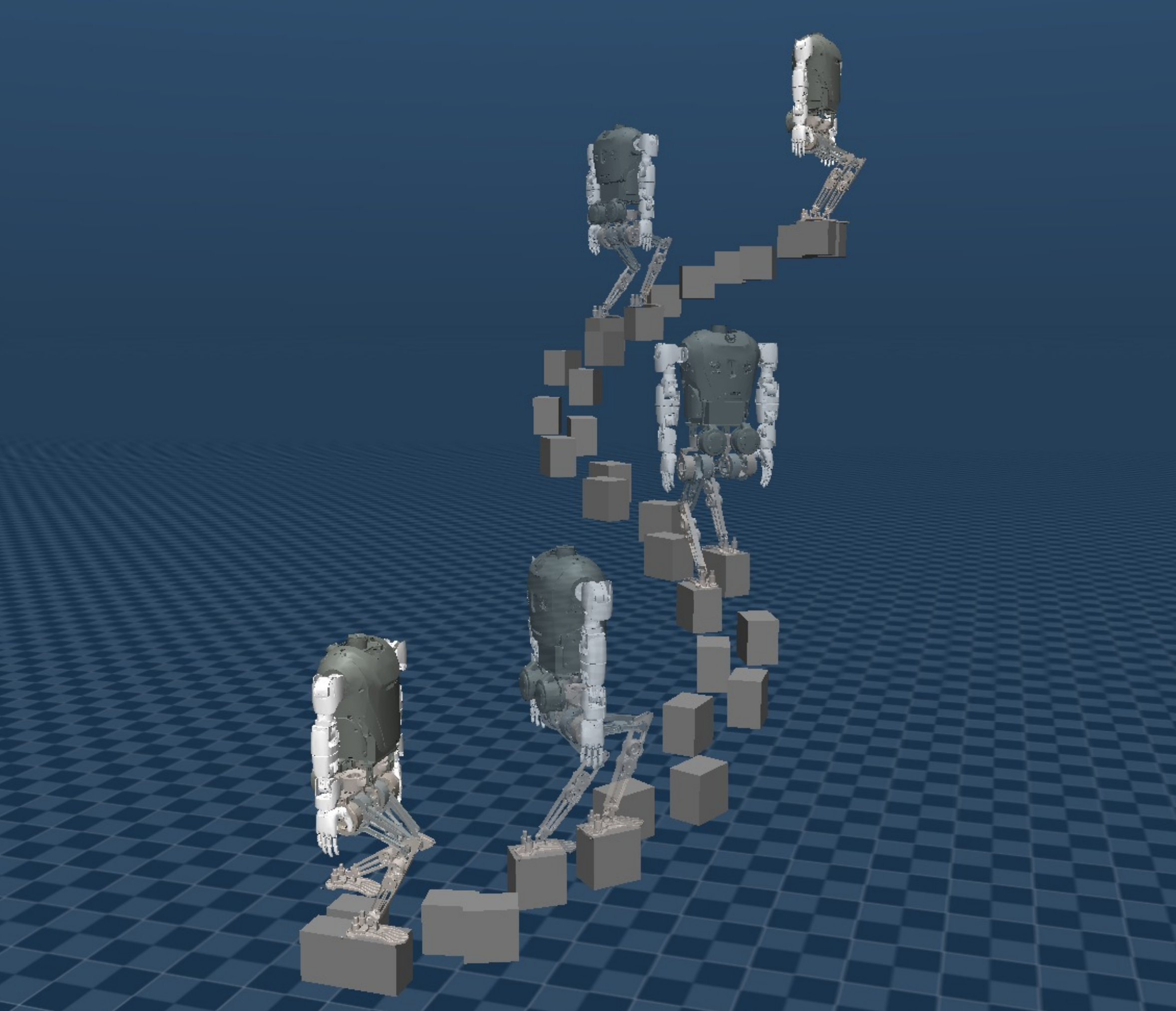}
    \caption{KUAVO traverses over randomly arranged 3D stepping stones, where adjacent stones have a rotation angle of 0.2 radians around the $z$-axis. The maximum random displacement amplitudes in the 3D space are [0.05, 0.05, 0.10] m along the $x$, $y$, and $z$ axes respectively.}
    \label{fig:upstair}
\end{figure}

\begin{figure}[htbp]
    \centering
    \includegraphics[width=1\linewidth]{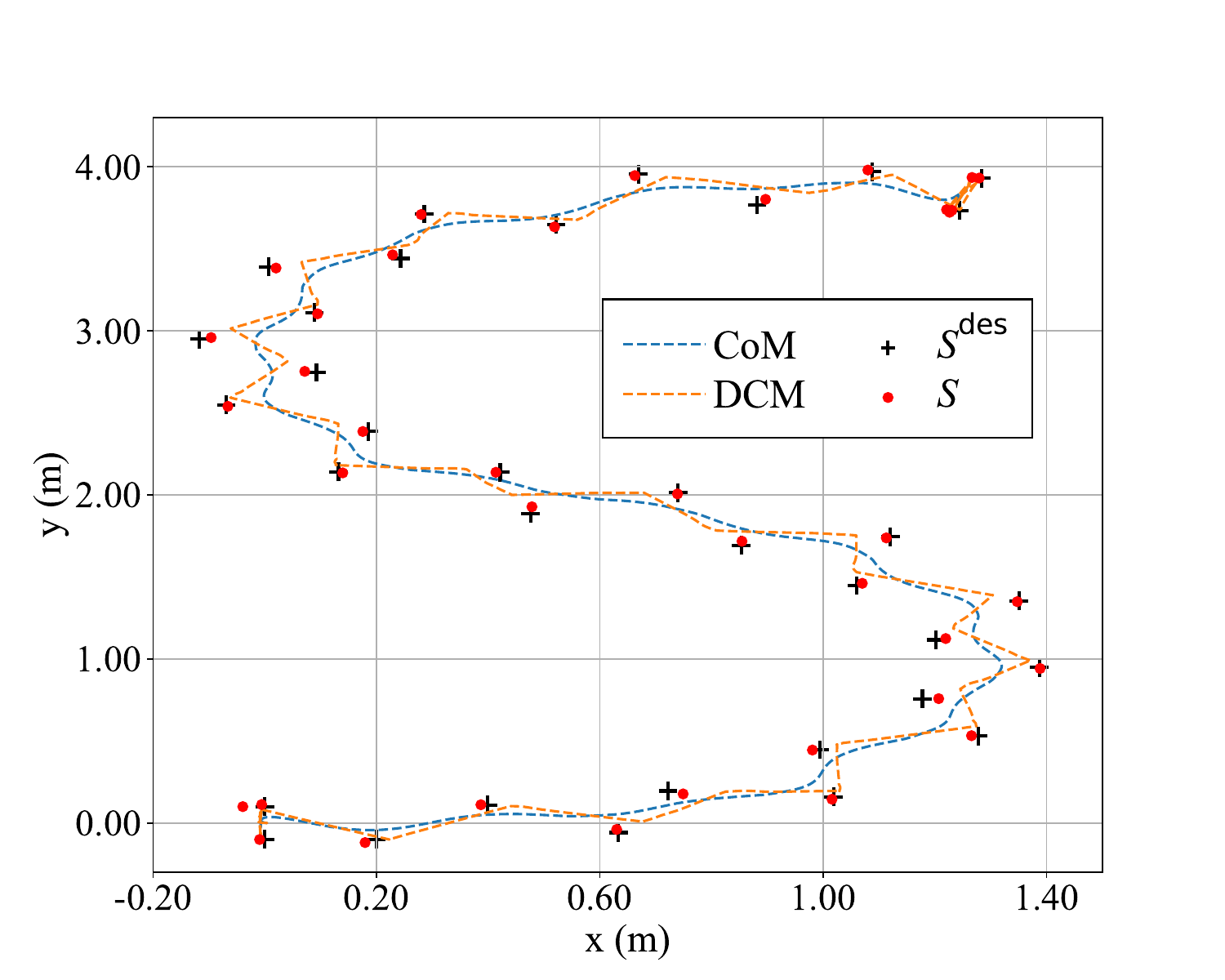}
    \caption{Profiles of the actual step positions (red circles) and their corresponding desired targets (black crosses) as KUAVO traverses randomly arranged 3D stepping stones. The red dashed line represents the DCM trajectory, while the blue dashed line represents the CoM trajectory.}
    \label{fig:multiple_curves1}
\end{figure}

\textbf{Scenario b:} In this scenario, the terrain was constructed with stepping stones arranged in a non-periodic three-dimensional configuration by relatively random placement in all directions. Additionally, the challenge complexity was enhanced through varied orientations and height differentials among the stepping stones, as depicted in Fig.~\ref{fig:upstair}. The KUAVO successfully travel this uneven terrain while maintaining an exceptionally low average position error $E_{avg}$ = \SI{0.017}{m}. Fig.~\ref{fig:multiple_curves1} comparatively shows three critical datasets: the actual step positions ($S$), theoretical target positions corresponding to stepping stone centroids ($S^{\mathrm{des}}$), and the corresponding trajectories of both the CoM and DCM.
% 增加discussion

\begin{figure}[h]
    \centering
    \includegraphics[width=1\linewidth]{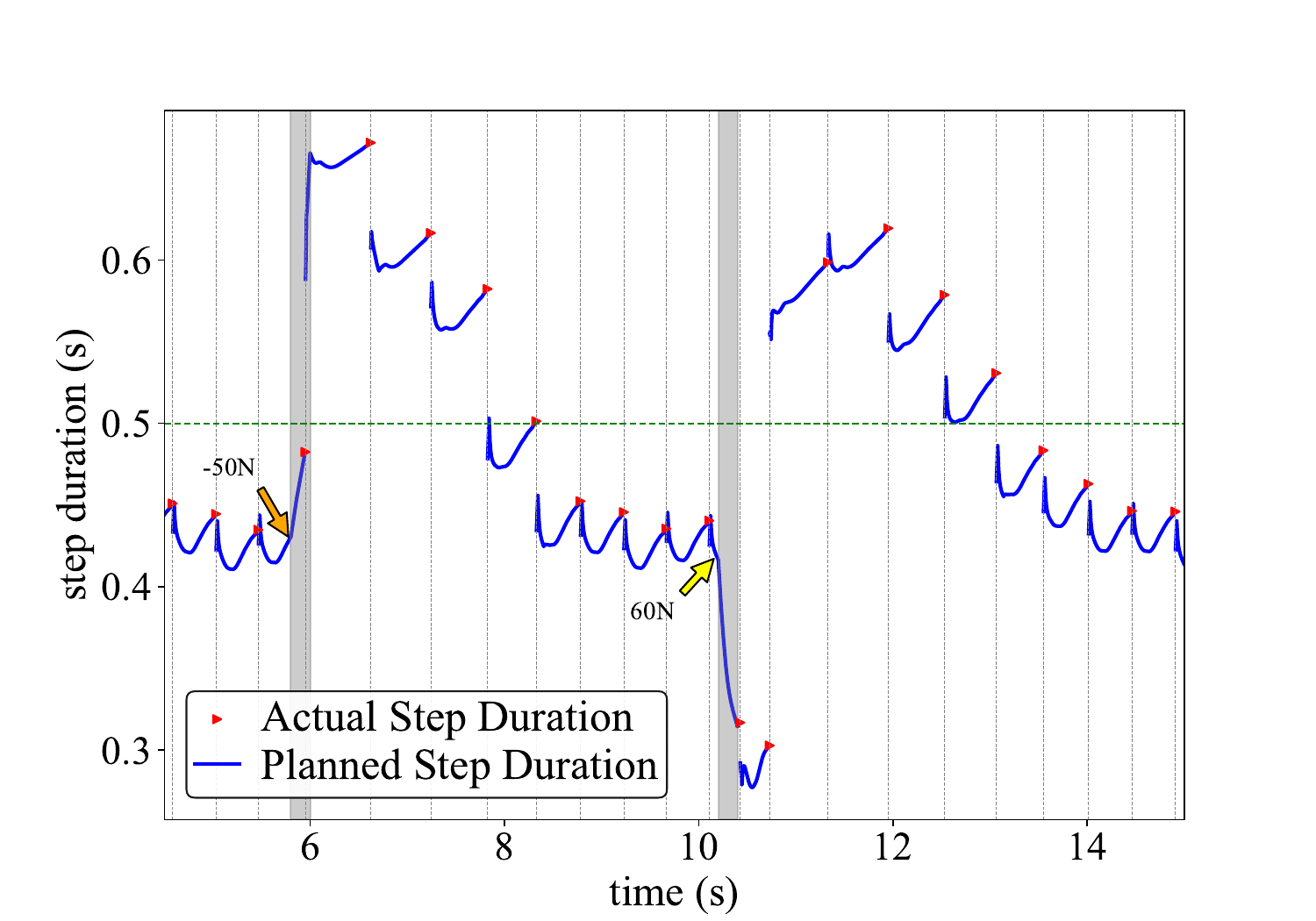}
    \caption{The MPC planner optimizes the step duration to maintain the robot's stability. The gray shaded area indicates the duration of the external forces implied. The vertical gray lines mark the touch-down moments of walking steps.}
    \label{fig:ts_disturbance}
\end{figure}

\textbf{Scenario c:} This scenario validated the robustness of the hierarchical control framework by applying several external forces during walking. 
% The experimental setup is shown in Table \ref{table:settings}, 
Unexpected external forces were applied in the sagittal plane: a backward force of \SI{50}{N} at around \SI{6}{s} and a forward force of \SI{60}{N} at around \SI{10}{s}. As shown in Fig. \ref{fig:ts_disturbance}, by adjusting the duration of the steps, the controller gradually brought the DCM back to normal values.

\begin{figure}[b]
    \centering
    \includegraphics[width=1\linewidth]{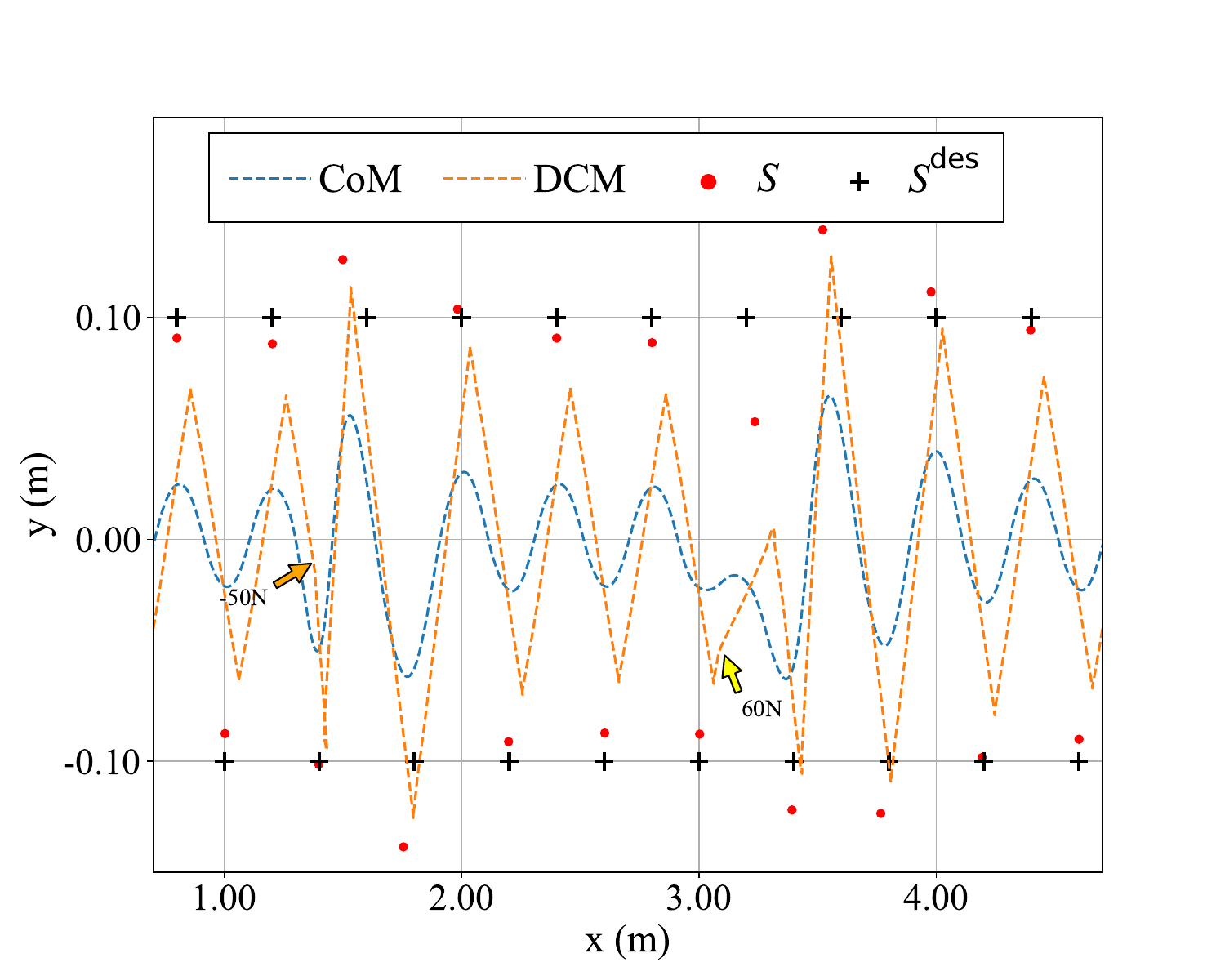}
    \caption{The evolution of the DCM and CoM when encountering external disturbances. The actual step position will significantly deviate from the desired position when subjected to external perturbations, and it will return to the normal position after a period of adjustment.}    
    \label{fig:dcm_disturbance}
\end{figure}

In the second validation, three experimental trials were compared on the same terrain without any disturbance of the terrain, where $\bm{p}^{\text{init}}=[20,0,0]$. The robot successfully traversed over the terrain with $\alpha=0$ and $\alpha=0.5$, but it failed at $36^{th}$ step with $\alpha=1$.

\begin{figure}[h]
    \centering
    \includegraphics[width=1\linewidth]{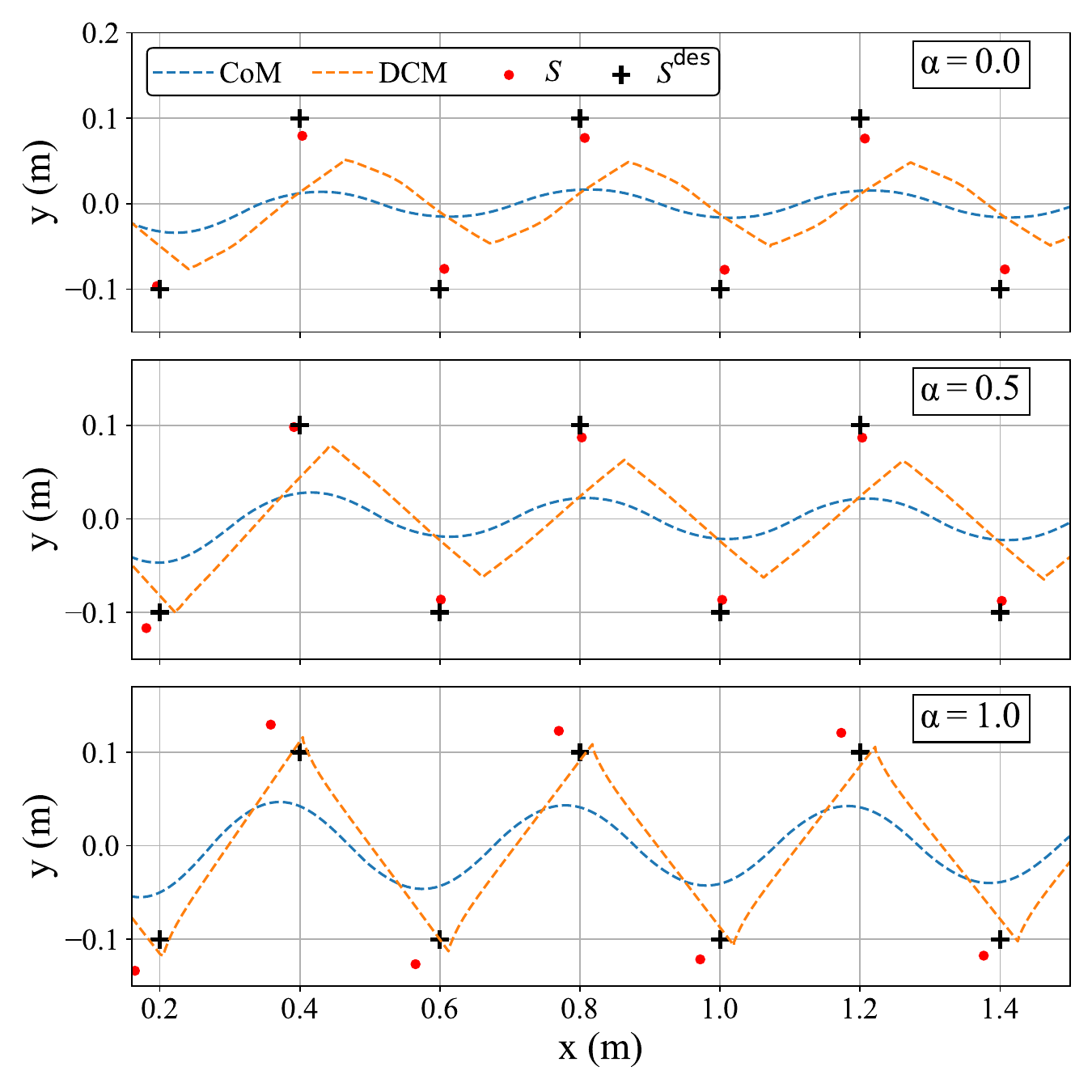}
    \caption{Profiles of the DCM and CoM for G-ALIP coefficients of $0$, $0.5$, and $1$. The actual step positions are closest to the desired positions when $\alpha = 0.5$.}
    \label{fig:alip_comp}
\end{figure}

To quantify planning performance, we introduced the average step position deviation metric $E_{avg}$, representing the average deviation between the actual and desired step positions. 
% As quantified in Table \ref{table:e_avg},
The $E_{avg}$ value observed at $\alpha=0.5$ ($E_{avg} = \text{\SI{0.012}{m}}$) is smaller than that at $\alpha=0$ ($E_{avg} = \text{\SI{0.023}{m}}$).
Fig. \ref{fig:alip_comp} illustrates the DCM variation curves during the process, indicating that the DCM and the CoM control are optimal at $\alpha=0.5$.

\begin{figure}[h]
    \centering
    \includegraphics[width=1\linewidth]{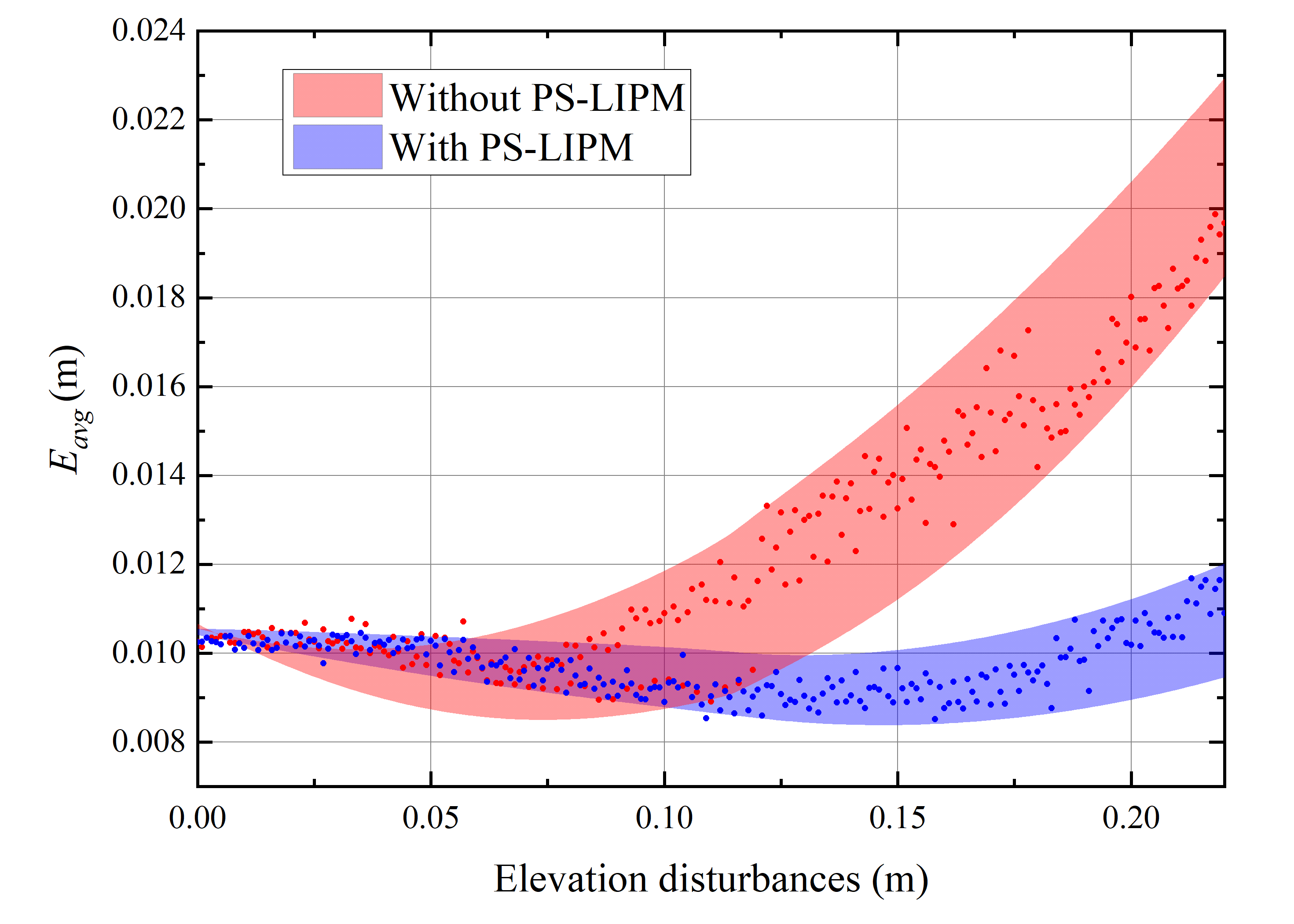}
    \caption{The average step position deviation $E_{avg}$ comparison profiles under terrain unevenness increases. The red-shaded area corresponds to conventional LIPM (without PS-LIPM) implementations, contrasting with PS-LIPM's performance profile in blue.}
    \label{fig:pslip_comp}
\end{figure}

The third validation is to evaluate the superiority of PS-LIPM over conventional LIPM. 
We compared the average step position deviation $E_{avg}$ for walking on elevation-disturbed stepping stones with and without PS-LIPM, where $\bm{p}^{\text{init}}=[20,0,0]$ cm, $\theta = 0$ and terrain disturbance in $z$ direction is $U[-10,10]$ (cm). The comparison of $E_{avg}$ is depicted in Fig.~\ref{fig:pslip_comp}. Data clusters without PS-LIPM (conventional LIPM, red-shaded area) exhibit systematically higher positional errors compared to PS-LIPM implementations (blue-shaded area) across equivalent elevation disturbances.

As elevation disturbances exceed \SI{0.15}{m}, the deviation in the red-shaded area increases more sharply, indicating a more significant degradation in performance without PS-LIPM. In contrast, the blue-shaded area is narrower as terrain unevenness increases, suggesting that the robot's performance is more consistent with PS-LIPM. This improvement in accuracy and consistency highlights the effectiveness of PS-LIPM in enhancing robotic mobility in challenging environments.

These investigations confirm that the modified MPC-based optimization architecture paradigm demonstrates enhanced adaptability to uneven terrain scenarios compared with conventional implementations.

\section{Conclusions}
This work presents a hierarchical MPC-WBC framework for robust bipedal locomotion on uneven terrains, advancing theoretical and practical capabilities through two key innovations. First, the proposed PS-LIPM overcomes conventional LIPM limitations by dynamically adjusting the CoM height during single-step cycles to accommodate terrain elevation variations. Second, the G-ALIP model generalizes centroidal momentum regulation, enabling CoM velocity compensation through CAM adjustments. The derived DCM step-to-step dynamics explicitly characterizes step transitions and enables simultaneous optimization of step positions and durations in MPC. Experimental evaluations demonstrate the adaptability of three-dimensional rotational stepping stones under random positional disturbances (maximum amplitude: [0.05, 0.05, 0.10] m) and robustness against external disturbances up to \SI{60}{N}. Comparative studies further confirm that PS-LIPM and G-ALIP outperform baseline methods, as evidenced by a lower average step position deviation.

\addtolength{\textheight}{-0cm}   % This command serves to balance the column lengths

%%%%%%%%%%%%%%%%%%%%%%%%%%%%%%%%%%%%%%%%%%%%%%%%%%%%%%%%%%%%%%%%%%%%%%%%%%%%%%%%

%%%%%%%%%%%%%%%%%%%%%%%%%%%%%%%%%%%%%%%%%%%%%%%%%%%%%%%%%%%%%%%%%%%%%%%%%%%%%%%%

%%%%%%%%%%%%%%%%%%%%%%%%%%%%%%%%%%%%%%%%%%%%%%%%%%%%%%%%%%%%%%%%%%%%%%%%%%%%%%%%

\bibliographystyle{IEEEtran} % 选择引用样式
\bibliography{./ref.bib}   % 指定 .bib 文件名（无需后缀）

\end{document}